# LGSE: Lexically Grounded Subword Embedding Initialization for Low-Resource Language Adaptation

**Hailay Kidu Teklehaymanot, Dren Fazlija, Wolfgang Nejdl**
L3S Research Center
Hannover, Germany
{teklehaymanot, dren.fazlija, nejdl}@L3S.de

## Abstract

Adapting pretrained language models to low-resource, morphologically rich languages remains a significant challenge. Existing vocabulary expansion methods typically rely on arbitrarily segmented subword units, resulting in fragmented lexical representations and loss of critical morphological information. To address this limitation, we propose the Lexically Grounded Subword Embedding Initialization (LGSE) framework, which introduces morphologically informed segmentation for initializing embeddings of novel tokens. Instead of using random vectors or arbitrary subwords, LGSE decomposes words into their constituent morphemes and constructs semantically coherent embeddings by averaging pretrained subword or FastText-based morpheme representations. When a token cannot be segmented into meaningful morphemes, its embedding is constructed using character n-gram representations to capture structural information. During Language-Adaptive Pretraining, we apply a regularization term that penalizes large deviations of newly introduced embeddings from their initialized values, preserving alignment with the original pretrained embedding space while enabling adaptation to the target language. To isolate the effect of initialization, we retain the original pre-trained model vocabulary and tokenizer and update only the new embeddings during adaptation. We evaluate LGSE on three NLP tasks: Question Answering, Named Entity Recognition, and Text Classification, in two morphologically rich, low-resource languages: Amharic and Tigrinya, where morphological segmentation resources are available. Experimental results show that LGSE consistently outperforms baseline methods across all tasks, demonstrating the effectiveness of morphologically grounded embedding initialization for improving representation quality in underrepresented languages. Project resources are available[1].

## 1. Introduction

Pretrained multilingual language models (PLMs) have become foundational in modern natural language processing (NLP), leveraging token sequences generated from word or subword-level units (Liu et al., 2024). A representative example is XLM-R (Conneau et al., 2020), a transformer-based model trained on over 100 languages using the SentencePiece algorithm for subword segmentation. Although XLM-R employs a shared vocabulary of 250K subword units, the effective average token coverage per language is relatively limited, approximately 2.5K subwords compared to monolingual models such as GPT, which typically utilize vocabularies in the range of 40K tokens (Wang et al., 2019).

Despite their wide language coverage, PLMs tend to favor high-resource languages, especially those that are typologically or orthographically closer to English (e.g., French, Spanish). In contrast, morphologically rich languages such as German face heightened out-of-vocabulary (OOV) challenges due to their complex inflectional and derivational systems (Ataman and Federico, 2018; Lample et al., 2018; Wang et al., 2019). These issues are significantly exacerbated for low-resource languages, particularly those writ-

Amharic Word for "egg": 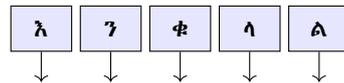

Subword-Based Embedding Initialization (BPE):

እ | ን | ቁ | ላ | ል

Embedding Composition:

E1, E2, E3, E4, E5 ⇒ Combined → $\vec{V}_{\text{noise}}$

Lexically Grounded Subword Embedding (LGSE):

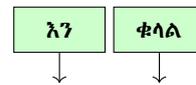

Embedding Composition:

E_ , E_ ⇒ Combined → $\vec{V}_{\text{meaningful}}$

Figure 1: Comparison of embedding initialization strategies: standard BPE subword splits vs. linguistically grounded morphemes.

ten in non-Latin scripts. Languages based on the Ge'ez script, such as Amharic and Tigrinya, suffer from poor lexical coverage and unreliable token representations due to a combination of script-specific orthographic complexity and minimal training data. To mitigate out-of-vocabulary (OOV) issues, subword tokenization techniques such as Byte Pair Encoding (BPE), introduced by Sennrich et al. (2016), have become foundational in neural machine translation (NMT) and broader NLP pipelines (Hiraoka et al., 2019; Bostrom and Durrett, 2020). However, BPE operates purely on character co-occurrence frequency and disregards linguistic structure, often fragmenting morphologically rich words into arbitrary subword units. This segmentation undermines semantic coherence, particularly in agglutinative or templatic languages. The issue is especially pronounced in underrepresented languages using the Ge'ez script, such as Amharic and Tigrinya, where BPE frequently breaks full lexical units, including complete nouns, into semantically meaningless fragments.

As illustrated in Figure 1, the Amharic word እንቁላል ('lənqulāl', meaning "egg") is decomposed into a series of subwords that fail to preserve its morphemic integrity. This fragmentation negatively affects subword embedding initialization by associating these noisy segments with ill-grounded or diluted vector representations, yielding embeddings that poorly capture the word's meaning. Consequently, morphologically aware tokenization is essential not merely for better segmentation but as a prerequisite for reliable and linguistically grounded embedding initialization in morphologically complex, low-resource languages.

Multilingual models like mBERT and XLM-R, which rely on shared vocabularies and embedding spaces across languages, often fail to encode the morphosyntactic nuances of such languages (Ahia et al., 2023; Wang et al., 2019). While language-adaptive pretraining (LAPT) (Chau et al., 2020) and similar transfer learning techniques aim to bridge this representational gap, they still struggle when applied to typologically distinct scripts. In particular, expanding the vocabulary or retraining the embedding matrix with newly introduced tokens disrupts alignment with the pretrained distribution, complicating the integration of linguistically informed tokenizers (Dobler and de Melo, 2023; de Vries and Nissim, 2021).

Although methods like vocabulary expansion and random or averaged embedding initialization provide partial relief, they fail to restore the structural grounding that morpheme-level units provide. This is especially critical for morphologically rich languages, where tokenization and embedding decisions are tightly coupled. As highlighted by Mofijul Islam et al. (2022) and Limisiewicz et al. (2023), subword-based models often produce semantically fragmented and unstable representations across languages, particularly in low-resource settings.

To address these limitations, we advocate for embedding strategies that align with morphologically aware tokenization. By respecting linguistic structure during both tokenization and embedding initialization, such methods promise not only improved representation quality but also fairer and more effective inclusion of underrepresented languages in NLP systems (Hangya et al., 2023; Teklehaymanot et al., 2025b). Our contributions are:

**(1)** We reveal that subword-based embeddings used in current multilingual pretrained models fail to capture the morphological structure of low-resource, morphologically rich languages, leading to fragmented and semantically weak representations;

**(2)** We propose Lexically Grounded Subword Embedding Initialization (LGSE), a novel strategy that respects linguistic boundaries by leveraging morpheme-aware segmentation for embedding initialization. Unlike conventional methods that rely on arbitrary subword fragments, LGSE creates semantically coherent representations, enabling more accurate and robust representation learning for underrepresented, morphologically rich languages such as Amharic and Tigrinya.

**(3)** We introduce the first human-annotated benchmark dataset for evaluating downstream NLP tasks and assessing model performance in identifying high-quality educational content for two morphologically rich, underrepresented languages, Amharic and Tigrinya. This publicly available resource[1] fills a critical gap for low-resource languages and provides a foundation for future research on cross-lingual transfer, morphological modeling, and educational AI.

**(4)** We rigorously evaluate LGSE on three downstream NLP tasks: Question Answering, Named Entity Recognition, and Text Classification, using two morphologically complex, low-resource languages of Amharic and Tigrinya. Compared to strong multilingual baselines, LGSE achieves substantial and consistent improvements over conventional subword-based embedding methods, demonstrating the effectiveness of linguistically grounded initialization in challenging language settings.

## 2. Related Work

### 2.1. Subword-Based Tokenization in Low-Resource Languages

Subword tokenization methods such as Byte-Pair Encoding (BPE) and SentencePiece are widely

---
[1] https://hailaykidu.github.io/LGSE-Project-/

used in multilingual pretrained language models. However, these approaches often cause excessive fragmentation when applied to morphologically rich and low-resource languages (Rust et al., 2021; Muller et al., 2021; Teklehaymanot and Nejdl, 2025). This over-segmentation leads to longer token sequences, which increase inference time (Hofmann et al., 2022; Sun et al., 2023), raise API costs (Ahia et al., 2023; Petrov et al., 2023), and degrade downstream task performance (Bostrom and Durrett, 2020; Fujii et al., 2023). Tokenizers trained on high-resource languages often produce segmentation mismatches in low-resource languages due to their lack of morphological awareness (Sun et al., 2023).

## 2.2. Morphologically Aware Tokenization

Morphologically aware tokenization addresses the limitations of conventional subword methods, which often ignore morpheme boundaries, inflating token counts and degrading performance on richly inflected languages. Approaches such as MorphBPE (Asgari et al., 2025) and MorphPiece (Jabbar, 2023) integrate linguistic structure with BPE, preserving morphemes and improving downstream accuracy across multiple languages. Korean morpheme-aware tokenizers enhance syntactic performance (Park et al., 2020), while MORSED (Goot et al., 2025) and Turkish hybrid tokenizers (Bayram et al., 2025) outperform BPE without sacrificing compression or semantic fidelity. MoVoC (Teklehaymanot et al., 2025a) demonstrates gains for low-resource Ge'ez-script languages. Token fertility analyses reveal a systematic "token tax" on morphologically complex languages, underscoring the efficiency and equity advantages of morphology-aware tokenization.

## 2.3. Vocabulary Expansion and Embedding Initialization

Vocabulary expansion is a common strategy for adapting pretrained models to underrepresented languages, particularly when the base vocabulary lacks coverage for non-Latin scripts or language-specific structures (Conneau et al., 2020; Downey et al., 2024; Pfeiffer et al., 2020). Pretrained models typically use a fixed vocabulary of approximately 50K tokens (Ushio et al., 2023), which often fails to represent morphologically rich or low-resource languages adequately.

To address out-of-vocabulary (OOV) issues, embedding initialization methods aim to leverage pretrained representations. Liu et al. (2021) propose synthesizing OOV embeddings using subword and hyperword information. UniBridge aligns non-overlapping tokens across languages via syntactic and semantic embeddings to enhance cross-lingual transfer (Pham et al., 2024). EVALM mitigates overfitting by initializing new tokens with high-resource language translations (InitHRL) and applying regularization during fine-tuning (Nag et al., 2023).

Recent work has also explored vocabulary expansion in decoder-only models such as LLaMA 2 and 3 to improve generative performance in low-resource settings (Balachandran, 2023; Larcher et al., 2023; Lin et al., 2024; Cui et al., 2023; Fujii et al., 2024; Choi et al., 2024b; Nag et al., 2025). These approaches typically add subwords based on frequency and continue pretraining or fine-tune with instruction data. While effective in reducing token overhead and improving fluency, they often rely on naive subword addition and initialization methods that do not account for linguistic structure. For example, Balachandran (2023) and Cui et al. (2023) introduce additional tokens for Tamil and Chinese using simple initialization, while Fujii et al. (2024) adapts LLaMA 2 for Japanese through cross-lingual pretraining. Similarly, Choi et al. (2024a) and Nguyen et al. (2024) extend coverage for Korean and Southeast Asian languages. However, these methods do not explicitly address fragmentation or incorporate morphological alignment in their tokenization strategies.

## 2.4. Linguistically Informed Embedding Alignment

Several studies have explored embedding reinitialization and alignment strategies that incorporate cross-lingual semantics. WECHSEL (Minixhofer et al., 2022) maps new subword embeddings to semantically similar words using multilingual vector alignment. Although it improves zero-shot transfer, it treats subwords as atomic units and overlooks morphological structure. OFA (Liu et al., 2024) introduces matrix factorization to compress the embedding space for scalable adaptation, yet it also ignores language-internal patterns. Language-specific vocabulary augmentation has been shown to improve syntactic tasks in low-resource languages (Chau et al., 2020), and Mundra et al. (2024) provides a comparative analysis of embedding initialization methods. Nonetheless, existing approaches largely neglect morpheme-based segmentation and do not exploit morphological composition for embedding initialization.

## 3. Problem Statement

Multilingual pretrained models such as mBERT and XLM-R employ a shared subword vocabulary $\mathcal{V}$ across multiple languages $\mathcal{L} = \{L_1, L_2, \ldots, L_m\}$. For a word $w \in L_i$, a tokenizer $T$ segments it into subwords $T(w) = [s_1, s_2, \ldots, s_n]$, where each sub-word $s_i \in \mathcal{V}$ is associated with a pretrained embedding $\mathbf{e}_i \in \mathbb{R}^d$. However, this subword segmentation frequently fails to align with the word's true morphemic structure $M(w) = [m_1, m_2, \ldots, m_k]$, where each $m_j$ represents a linguistically meaningful morpheme.

This misalignment is particularly problematic for morphologically rich and low-resource languages, leading to suboptimal semantic representations and poorer generalization on unseen or infrequent tokens.

## 4. Vocabulary Expansion and Initialization

This section introduces two key approaches to enhancing embedding initialization for morphologically rich and low-resource languages. Section 4.1 presents Lexically Grounded Subword Embedding Initialization Framework (LGSE) which leverages subword-level semantic representations from FastText (Bojanowski et al., 2017) to initialize embeddings for morpheme-aligned tokens. This method ensures that the initialized vectors capture meaningful morphological and semantic patterns, aligning with the language's internal structure. In Section 4.2, we describe embedding initialization for new morphologically grounded tokens, which addresses out-of-vocabulary (OOV) scenarios by generating embeddings for novel morpheme-based units using composition strategies informed by morphological structure and distributional semantics. Together, these strategies aim to improve vocabulary coverage, semantic coherence, and representation quality in low-resource, morphologically complex languages.

### 4.1. Lexically Grounded Subword Embedding Initialization (LGSE)

Given access to a morphologically-aware tokenizer and pretrained FastText embeddings, we represent a new token $t$ segmented into morphemes $M(t) = [m_1, m_2, \ldots, m_k]$. Each morpheme $m_j$ is further represented by a set of character $n$-grams

$$G_j = \{g_{j1}, g_{j2}, \ldots, g_{jn_j}\}.$$

Each $n$-gram $g$ has an associated FastText embedding $\mathbf{f}_g \in \mathbb{R}^d$. The embedding for morpheme $m_j$ is computed as the average of its constituent $n$-gram embeddings

$$\mathbf{m}_j = \frac{1}{|G_j|} \sum_{g \in G_j} \mathbf{f}_g$$

while the initial token embedding $\mathbf{e}_t$ is obtained by averaging over all morpheme embeddings, i.e.,

$$\mathbf{e}_t = \frac{1}{k} \sum_{j=1}^{k} \mathbf{m}_j = \frac{1}{k} \sum_{j=1}^{k} \left( \frac{1}{|G_j|} \sum_{g \in G_j} \mathbf{f}_g \right).$$

To align the FastText embedding space with the pretrained model embedding space, a learned linear projection $\mathbf{W} \in \mathbb{R}^{d \times d}$ is applied, i.e.,

$$\mathbf{e}_t^{\text{aligned}} = \mathbf{W} \mathbf{e}_t.$$

### 4.2. Embedding Initialization for New Lexically Grounded Subword Tokens

We initialize the embedding for a new token as the average of its morpheme embeddings computed via FastText-based pooling:

$$\mathbf{e}_{\text{new}} = \frac{1}{k} \sum_{j=1}^{k} \mathbf{m}_j^{\text{aligned}},$$

where $\mathbf{m}_j^{\text{aligned}}$ are morpheme embeddings after projection. For tokens without known morpheme segmentation or embeddings, we initialize by sampling from a multivariate normal distribution estimated from existing pretrained embeddings:

$$\mathbf{e}_{\text{new}} \sim \mathcal{N}(\mu, \Sigma),$$

where $\mu$ and $\Sigma$ are the mean and covariance matrix of pretrained embeddings. To prevent excessive deviation of new embeddings from their initialization during continual pretraining or fine-tuning, we apply the regularization loss

$$\mathcal{L}_{\text{reg}} = \lambda \left\| \mathbf{e}_{\text{new}} - \boldsymbol{\mu} \right\|^2,$$

where $\boldsymbol{\mu}$ is the initial embedding vector (e.g., from FastText projection), and $\lambda$ controls the regularization strength, balancing stability and adaptability.

## 5. Language-Adaptive Pretraining (LAPT)

To enhance the XLM-R model's performance on morphologically complex, low-resource languages such as Tigrinya and Amharic, we move away from subword-based approaches that utilize BPE vocabularies, such as FOCUS (Dobler and de Melo, 2023). Instead, we initialize the embedding layer with lexically grounded representations derived from a morphology-aware tokenizer trained on linguistically annotated corpora. This tokenizer segments text into morphemes, preserving the language's meaningful lexical and grammatical structures, unlike arbitrary subword units.

We employ *Language-Adaptive Pretraining (LAPT)* with a morpheme-level Masked Language

Modeling (MLM) objective, initializing the embedding layer with morpheme-aware representations from annotated corpora to maintain the linguistic integrity of the target languages. For Amharic, we utilize the CC100 corpus (133M tokens), previously used in XLM-R pretraining (Conneau et al., 2020), while for Tigrinya, we rely on data from (Gaim et al., 2021), totaling approximately 0.5GB. Hyperparameters for both languages are consistent, as detailed in Table 1.

We preserve the pretrained XLM-R encoder parameters $\{L_1, L_2, \ldots, L_n\}$ and adapt only the embedding layer $E$, initializing it with a language-specific vocabulary $\mathcal{V}_{\text{morph}}$ tailored to each target language. Pretraining is performed on monolingual Tigrinya and Amharic corpora, applying a dynamic masking probability of 15% to sequences that are either truncated or padded to a maximum length of 256 tokens.

## 6. Experimental Setup

Our experiments are conducted using the multilingual encoder-based model XLM-R (Conneau et al., 2020) as the foundational architecture. XLM-R is selected for its proven cross-lingual transfer performance and extensive use in multilingual NLP research. Its decoupled SentencePiece tokenizer enables straightforward integration of morpheme-level tokens without modifying the underlying model architecture. To ensure fair comparison and reproducibility, all experiments utilize the base version of XLM-R and maintain consistent hyperparameters across both baseline and LGSE-enhanced models.

The model contains approximately 125 million parameters. Training was performed on a single GPU with 4 CPU cores and 46 GB RAM, with each run allocated up to 24 GPU hours on an Ampere architecture GPU. The computational environment was managed using Anaconda to ensure consistency and reproducibility.

### 6.1. Linguistically Informed Hybrid Tokenization

We adopt a **morphologically informed tokenization strategy** proposed in MoVoC (Teklehaymanot et al., 2025a) for segmenting words into lexically grounded morphemes using supervised morphological analysis applied to monolingual corpora $P_{\text{am}}$ (Amharic) and $P_{\text{ti}}$ (Tigrinya). Unlike conventional tokenizers that rely solely on frequency-based subword segmentation, the approach respects linguistic boundaries to preserve morphological integrity. To construct a vocabulary that is both *linguistically meaningful* and *computationally efficient*, we combine high-frequency morpheme tokens with subword units learned via **Byte-Pair Encoding (BPE)**. A hyperparameter $r \in [0, 1]$ controls the ratio of morpheme tokens, yielding a hybrid vocabulary:

$$\begin{aligned} V &= V_{\text{BPE}_{\text{small}}} \cup V_{\text{morph}}, \\ |V_{\text{BPE}_{\text{small}}}| &= s(1-r), \\ |V_{\text{morph}}| &= sr. \end{aligned} \quad (1)$$

Table 1: Hyperparameter settings used for further pretraining with morpheme-aware tokenization and fine-tuning.

| Hyperparameter | Value |
| --- | --- |
| Maximum sequence length | 256 |
| Batch size | 32 |
| Number of training epochs | 10 |
| Learning rate | $5 \times 10^{-5}$ |
| Learning rate schedule | Constant |
| MLM probability | 0.15 |
| Weight decay | 0.01 |
| Optimizer | Adam |
| Adam $\epsilon$ | $1 \times 10^{-8}$ |
| Adam $\beta_1$ | 0.9 |
| Adam $\beta_2$ | 0.999 |
| Mixed precision (fp16) | True |

Tokenization proceeds in two stages: $(i)$ words are first segmented into morphemes; $(ii)$ BPE is then applied **within each morpheme**, preventing merges across morpheme boundaries.

Formally, for a word $w = m_1 m_2 \cdots m_k$, the tokenizer output is:

$$\text{Tokenizer}(w) = \bigcup_{i=1}^{k} \text{BPE}_{\text{small}}(m_i).$$

This **morphology-aware BPE** forms the foundation of our **Lexically Grounded Subword Embedding Initialization (LGSE)** framework. By aligning embeddings with linguistically interpretable morphemes and subwords, LGSE mitigates semantic fragmentation and noise introduced by arbitrary subword splits, thereby producing representations that better capture the morphological richness of underrepresented languages. For practical efficiency, we apply the pre-trained MoVoC (Teklehaymanot et al., 2025a) tokenizer to our parallel and monolingual corpora from the **No Language Left Behind (NLLB)** project (Fan et al., 2021) for both Amharic and Tigrinya, generating tokenized sequences with morphologically-informed subword units.

### 6.2. Baselines

To evaluate the effectiveness of our proposed LGSE approach, we compare it against several strong baselines. In all cases, the original XLM-R encoder layers remain frozen during initialization to isolate the effect of embedding strategies. All models subsequently undergo Language Adaptive Pretraining (LAPT) under identical settings for fairness.

- **XLM-R Off-the-Shelf:** The unmodified XLM-R model is used in a zero-shot setting without any additional training. This baseline provides a reference point for assessing the inherent transfer capabilities of the pretrained model in our target languages.

- **XLM-R + LAPT:** The original XLM-R vocabulary and embeddings are preserved, and the model is further adapted using Language Adaptive Pretraining on monolingual target language data. This measures the gains from LAPT alone without modifying the tokenizer or embeddings.

- **Random Initialization for Newly Added Tokens + LAPT:** When expanding the vocabulary with morphologically grounded tokens, only the embeddings for these new tokens are randomly initialized, while the pretrained embeddings and encoder parameters remain unchanged. Each new embedding vector is sampled from a Gaussian distribution estimated from the original embedding matrix:

$$\mathbf{e}_t \sim \mathcal{N}(\mu, \Sigma),$$

  where $\mu$ and $\Sigma$ are the empirical mean and covariance of the original embeddings. This baseline isolates the contribution of linguistically informed initialization by comparing against a purely random, statistically coherent initialization strategy.

- **Subword-Based Initialization (FOCUS) + LAPT:** We adopt FOCUS (Dobler and de Melo, 2023), a subword-level embedding refinement method that computes weighted combinations of overlapping pretrained subword tokens using Sparsemax. This improves representations for rare or unseen tokens without modifying the original tokenizer or vocabulary.

- **Lexically Grounded Subword Embedding Initialization (LGSE) + LAPT:** Our proposed approach combines morphology-aware tokenization with embedding initialization based on FastText-derived morpheme embeddings, aligned via a learned projection layer. This linguistically informed strategy mitigates over-fragmentation and enhances coverage of morphologically rich words, improving representation quality for low-resource languages.

## 7. Evaluation

We evaluate our morphology-aware tokenization by comparing tokenization efficiency, token fertility, and inference latency against standard BPE for Amharic and Tigrinya. Furthermore, we assess our proposed models on a range of downstream NLP tasks across these two morphologically rich, low-resource languages that use the Ge'ez script : **Amharic** and **Tigrinya**. These languages were selected due to the availability of supervised, morphologically annotated data as well as curated evaluation datasets. We conduct experiments on three key tasks: text classification, question answering, and named entity recognition. The hyperparameters used for all evaluation tasks are provided in Table 1.

**Text Classification:** We address the task of assigning predefined labels to input texts, with a specific focus on evaluating the quality of educational content.

**Educational Quality Classification Dataset:** To support this task, we introduce a new benchmark dataset comprising 2,500 human-annotated samples in **Amharic** and **Tigrinya**. The dataset was developed in close collaboration with local linguistic communities to ensure cultural and linguistic relevance. Data collection proceeded in two stages: initially, a diverse set of texts was sourced from publicly available educational materials, including manuals and blog posts; subsequently, each text was annotated on a 1-6 scale reflecting perceived educational quality. Comprehensive dataset statistics and illustrative examples will be provided in the Appendices in the camera-ready version. For model training and evaluation, the dataset was carefully curated and split into 80% for training, 10% for development, and 10% for testing.

**Named Entity Recognition (NER):** We perform NER experiments using the balanced train-dev-test splits of the **MasakhaNER** dataset (Adelani et al., 2021) for Amharic and for the **Tigrinya NER dataset** (Yohannes and Amagasa, 2022), where no official data split is provided, we create a consistent partition by randomly splitting the data into **80%** for training, **10%** for development, and **10%** for testing. Model selection is based on performance on the development set, and final results are reported on the test set.

**Question Answering (QA):** QA performance is evaluated on the **TIGQA** train-dev-test splits balanced dataset (Teklehaymanot et al., 2024), which contains expert-annotated question-answer pairs in Tigrinya. For Amharic, we use the **AmQA**, train-dev-test splits dataset (Taffa et al., 2024), developed for low-resource QA benchmarking. The final results are reported on the test set for both

Table 2: Performance of XLM-R across three NLP tasks in Tigrinya and Amharic. F1 score is used for QA and NER; Accuracy is used for TC. All results are reported as mean ± standard deviation over five runs. The best performance per task is highlighted in bold.

| Model | Task Category | Task | Metric | Tigrinya | Amharic | Avg |
|---|---|---|---|---|---|---|
| XLM-R (off-the-shelf) | Question Answering | QA | F1 | 61.3 ± 0.4 | 71.4 ± 0.9 | 66.35 |
| | Text Classification | TC | AC | 63.2 ± 0.7 | 70.1 ± 0.6 | 66.65 |
| | Named Entity Recognition | NER | F1 | 66.4 ± 0.6 | 70.2 ± 0.8 | 68.3 |
| XLM-R + LAPT | Question Answering | QA | F1 | 70.5 ± 0.8 | 74.9 ± 0.5 | 72.7 |
| | Text Classification | TC | AC | 69.4 ± 0.5 | 71.0 ± 0.4 | 67.8 |
| | Named Entity Recognition | NER | F1 | 69.8 ± 0.5 | 75.0 ± 0.6 | 70.4 |
| XLM-R + Random + LAPT | Question Answering | QA | F1 | 68.7 ± 0.6 | 71.3 ± 0.8 | 70 |
| | Text Classification | TC | AC | 69.9 ± 0.6 | 70.8 ± 0.8 | 70.35 |
| | Named Entity Recognition | NER | F1 | 70.3 ± 0.7 | 74.0 ± 0.7 | 72.15 |
| XLM-R + FOCUS + LAPT | Question Answering | QA | F1 | 75.5 ± 0.3 | 77.8 ± 1.0 | 76.65 |
| | Text Classification | TC | AC | 72.4 ± 0.4 | 76.5 ± 0.9 | 74.45 |
| | Named Entity Recognition | NER | F1 | 77.5 ± 0.4 | 78.1 ± 0.9 | 77.8 |
| XLM-R + LGSE + LAPT | Question Answering | QA | F1 | **78.0 ± 0.4** | **78.5 ± 0.4** | **78.25** |
| | Text Classification | TC | AC | **75.2 ± 0.5** | **77.8 ± 0.3** | **76.5** |
| | Named Entity Recognition | NER | F1 | **79.0 ± 0.3** | **79.4 ± 0.4** | **79.2** |

Amharic and Tigriyna QA datasets. We report **F1 scores** for NER, QA, and Text classification. Each experiment is repeated **five times** with different random seeds. We report the **mean and standard deviation** of results. The complete training configurations and hyperparameter settings are presented in Table 1.

We compare our approach against the baselines mentioned in Section 6.2.

Unlike these baselines, our method (LGSE) explicitly incorporates **morpheme-level structure**, which we argue is essential for capturing the deep semantics of morphologically complex languages such as Amharic and Tigrinya.

## 8. Results and Discussion

The results in Table 2 demonstrate a clear and consistent performance improvement when applying our proposed methods across all three tasks to the XLM-R model.

### 8.1. Baseline Performance

The off-the-shelf XLM-R model yields the lowest performance across all tasks. This is expected, as the model has not been adapted to the specific languages or domains involved. For instance, it achieves an average QA F1 score of **66.35** and NER F1 score of **68.30**, indicating limited ability to generalize to Tigrinya and Amharic without further adaptation.

### 8.2. Impact of Language-Adaptive Pretraining (LAPT)

Applying **Language-Adaptive Pretraining (LAPT)** substantially improves performance across all tasks. QA and NER scores increase by approximately 6-7 percentage points on average, confirming the benefit of continued pretraining on language-specific data for low-resource scenarios.

### 8.3. Effect of Embedding Initialization Methods

Beyond LAPT, we examine the impact of different subword embedding initialization methods: Random, FOCUS, and our proposed Lexically Grounded Subword Embedding Initialization (LGSE). The **FOCUS + LAPT** configuration outperforms the **Random + LAPT** baseline, achieving a QA F1 score of **76.65** and NER F1 of **77.80**. This indicates that more informed subword representations can lead to better convergence and improved performance.

### 8.4. Effectiveness of LGSE and Cross-Language Impact

The proposed method, **LGSE + LAPT**, which integrates Language-Adaptive Pretraining with **Lexically Grounded Subword Embedding Initialization (LGSE)**, achieves the best overall performance, obtaining QA F1 of **78.25**, TC accuracy of **76.50**, and NER F1 of **79.20**. LGSE employs a **morpheme-aware tokenizer** that captures linguistically meaningful units, offering im-proved representations for morphologically rich and low-resource languages such as Amharic and Tigrinya. Unlike conventional subword-based approaches, this method aligns with the underlying morphological structure of these languages, thereby enhancing semantic fidelity and reducing segmentation errors.

Our analysis further reveals that **vocabulary overlap** plays a non-trivial role in cross-lingual embedding transfer. Despite Tigrinya's absence in pretraining corpora, we observe approximately 1,280 shared morphemes with Amharic, largely driven by code-mixing rather than strict linguistic similarity. While this overlap facilitates partial transfer, it also introduces potential semantic drift. To address rare and out-of-vocabulary morphemes, LGSE leverages **FastText-based character n-gram embeddings**, enabling compositional representations and robust initialization, which are crucial for improving generalization in low-resource settings.

**Cross-Language Impact.** Although Amharic benefits from relatively larger resources, LGSE substantially reduces the performance gap with Tigrinya. This improvement underscores the effectiveness of **linguistically informed tokenization and embedding strategies** in supporting cross-lingual generalization under severe resource constraints, particularly for morphologically complex languages.

### 8.5. Tokenization Metrics and Efficiency

As mentioned in Section 6.1, to evaluate the practical utility of our tokenization approach, we adopt the pre-trained MoVoC tokenizer (Teklehaymanot et al., 2025a) for Amharic and Tigrinya corpora from the **No Language Left Behind (NLLB)** project (Fan et al., 2021). To highlight the advantages of our approach, we provide illustrative tokenization examples comparing our morphology-aware method with conventional BPE. For instance, the Tigrinya sentence: "ሰላም ንኩሉፈትኡር" (selam nəkulufət'ur) is tokenized into 21 tokens using standard BPE, whereas our morphology-aware approach produces only 6 tokens, preserving morphemes and reducing over-segmentation. We define **Token Fertility (TF)** as

$$\text{TF} = \frac{\text{Total tokens}}{\text{Total words}}.$$

Lower TF indicates fewer redundant tokens per word.

Additionally, we measure **inference latency (IL)** across different computational budgets to evaluate efficiency trade-offs. Our analysis consistently shows that morphology-aware tokenization reduces sequence lengths, lowers TF, and decreases IL, demonstrating both computational efficiency and practical utility in morphologically rich, low-resource languages.

## 9. Conclusion

We propose a Lexically Grounded Subword Embedding Initialization (LGSE) framework for morphologically rich, low-resource languages, focusing on Amharic and Tigrinya. By combining morpheme-aware tokenization with FastText-based compositional embeddings and Language-Adaptive Pretraining (LAPT), LGSE consistently improves performance across multiple downstream tasks. These results underscore the benefits of incorporating lexical and morphological structure into multilingual NLP models.

## 10. Ethical Considerations and Limitations

**Limitations and Future Work.** While the proposed framework demonstrates promising improvements, it faces several limitations. First, it depends on morphologically annotated resources, which remain scarce for many low-resource languages, constraining its applicability in truly multilingual settings. Second, the current design targets encoder-based architectures such as XLM-R, limiting direct integration with decoder-based or sequence-to-sequence models widely used in machine translation and other generative tasks. Third, the incorporation of Lexically Grounded Subword Embedding Initialization introduces additional computational overhead compared to frequency-driven subword segmentation methods, which may impact scalability for very large vocabularies or low-resource deployment environments.

As future work, we plan to extend the framework to decoder-based and encoder–decoder architectures, enabling its use in machine translation and generative modeling. Additionally, we aim to investigate vocabulary *replacement* versus *expansion* strategies under these settings to better understand their trade-offs in terms of efficiency and performance across diverse language families.

**Ethical Considerations.** This work uses only publicly available datasets, with all sources properly cited to ensure transparency. ChatGPT was used only for paraphrasing and language clarity no scientific content was generated. The Amharic and Tigrinya annotated datasets, models, and code will be released under an open-access license to support research equity and inclusivity. No personally identifiable information (PII) or sensitive content is involved. All research activities adhere to established ethical guidelines for NLP, with attention to linguistic and cultural sensitivity in underrepresented language communities. Our goal is to promote responsible and inclusive cross-lingual NLP development.

## 11. Acknowledgements

This research was supported by the German Academic Exchange Service (DAAD) through the Hilde Domin Programme (funding no. 57615863).

## 12. Bibliographical References